\pdfoutput=1

\documentclass[11pt]{article}

\usepackage[]{acl}

\usepackage{times}
\usepackage{latexsym}

\usepackage[T1]{fontenc}

\usepackage[utf8]{inputenc}

\usepackage{microtype}

\usepackage{multirow}
\usepackage{graphicx}
\usepackage{amsmath}

\usepackage[markup=underlined]{changes}

\title{FreeTransfer-X: Safe and Label-Free Cross-Lingual Transfer from Off-the-Shelf Models}

\author{Yinpeng Guo \and Liangyou Li \and Xin Jiang \and Qun Liu \\
  Huawei Noah's Ark Lab \\
  \texttt{\{guo.yinpeng, liliangyou, jiang.xin, qun.liu\}@huawei.com}}

\begin{document}
\maketitle
\begin{abstract}
Cross-lingual transfer (CLT) is of various applications. 
However, labeled cross-lingual corpus is expensive or even inaccessible, especially in the fields where labels are private, such as diagnostic results of symptoms in medicine and user profiles in business.
Although being lack of labels, there are off-the-shelf models in these sensitive fields. 
Instead of pursuing the original labels, a workaround for CLT is to transfer knowledge from the off-the-shelf models without labels. 
To this end, we define a novel CLT problem named \textit{FreeTransfer-X} that aims to achieve knowledge transfer from the off-the-shelf models in rich-resource languages.
To address the problem, we propose a 2-step knowledge distillation~\citep[KD,][]{hinton2015distilling} framework based on multilingual pre-trained language models (mPLM)\footnote{Source code are available at \href{https://github.com/huawei-noah/noah-research/tree/master/NLP/FreeTransfer-X}{https://github.com/huawei-noah/noah-research/tree/master/NLP/FreeTransfer-X}}.
The significant improvement over strong neural machine translation (NMT) baselines demonstrates the effectiveness of the proposed method.
In addition to reducing annotation cost and protecting private labels, the proposed method is compatible with different networks and easy to be deployed.
Finally, a range of analyses indicate the great potential of the proposed method.

\end{abstract}

\begin{figure*}[!ht]
    \centering
    \includegraphics[width=1.\textwidth]{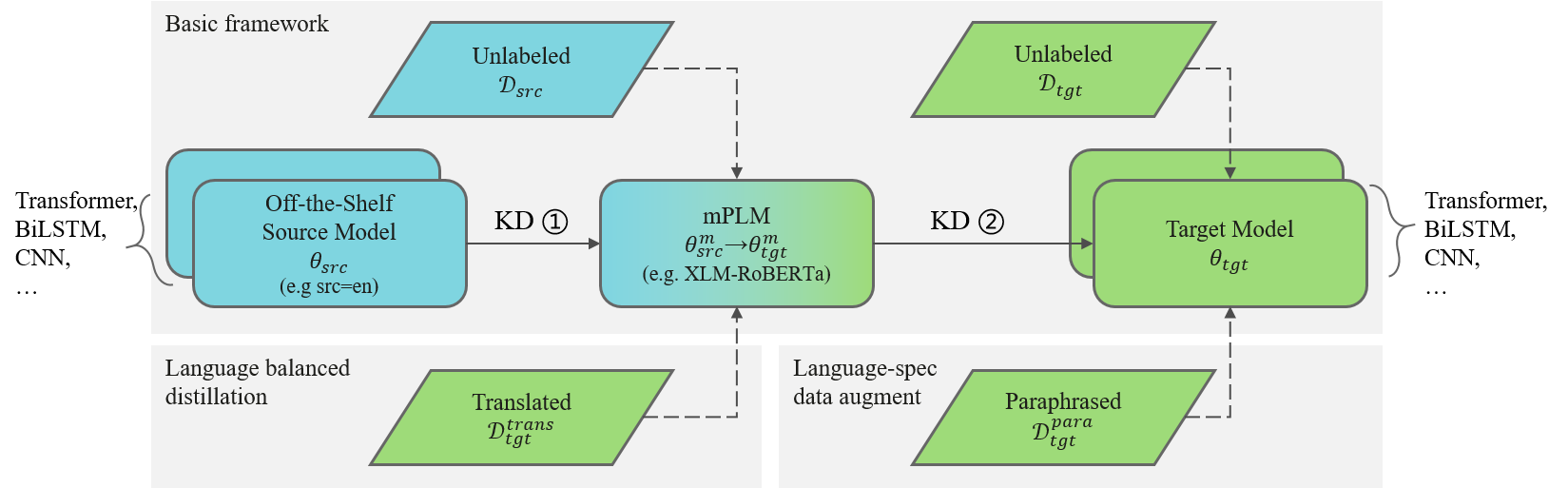}
    \caption{Overview of the proposed 2-step knowledge distillation (KD) framework. KD-(1) distills knowledge from the off-the-shelf English model to the mPLM. KD-(2) distills knowledge from the mPLM to the model in the target language. {\color{Turquoise}Blue modules}: in the source language $src$, {\color{LimeGreen}green modules}: in the target language $tgt$.}
    \label{fig:overview}
\end{figure*}

\section{Introduction}
Cross-lingual transfer (CLT) is a critical topic for natural language processing due to the data imbalance between languages. 
While models of rich-resource languages (e.g. English) have been applied on various real-world tasks, 
the progress on poor-resource languages lags behind.
CLT researches enable the knowledge transfer from the rich-resource languages to the poor-resource languages.

Although the application of CLT is valuable, data labels are expensive or even inaccessible in private and sensitive domains, such as medicine and business.
For example, the diagnostic results of a user's symptoms are private and a company's internal description of users are confidential.
Since short of labels for CLT, even though there are excellent applications in rich-resource languages, it is difficult to benefit the people using poor-resource languages.
Previous CLT researches have not well studied how to leverage knowledge of rich-resource languages without labels.
To define and tackle this problem will benefit both the community and the industry. 

In order to reduce the demand of labels, existing works mainly fall into two paradigms as follows.
One paradigm focuses on learning language-agnostic representation and model parameters. 
CLT is realized by either aligning parameters of monolingual models or sharing parameters among different languages~\citep{liu-etal-2019-investigating,jacob-mbert-2019,xlmr-conneau-etal-2020-unsupervised,Wang*2020Cross-lingual}. 
The objective is to build a unified representation, which is used by downstream tasks, for all the languages.
In this paradigm, although the demand of labels is reduced, it still requires a certain number of labels to adapt the model to a particular language and task.
Besides, models in this paradigm are usually large-scale Transformers~\citep{trm_NIPS2017_3f5ee243} based on mPLMs, which limits their deployment in real-world.
Another paradigm is to leverage machine translation (MT) systems to generate training or testing pseudo-corpus for a specific language~\cite{xnli-conneau-etal-2018}.
For simplicity, we take English as the rich-resource languages in this paper.
\texttt{Translate-train} translates annotated training corpus from English to other languages. Gold labels are directly applied to the translated data. Although labels in poor-resource languages are not required, gold labels in English are still necessary. 
On the contrary, \texttt{Translate-test} translates testing corpus from poor-resource languages to English. This method can directly leverage off-the-shelf English models, but it runs a 2-pass inference which highly limits its efficiency. 
Both the two CLT paradigms mentioned above require language-specific and task-specific labels, except for the 2-pass \texttt{Translate-test}. 
The demand of labels highly limits the reuse of the English knowledge in private and sensitive domains. 
Then a question comes up: \textit{Is it possible to perform CLT totally without labels?}


In this paper, we define a novel problem: \textit{safe and label-free cross-lingual transfer from off-the-shelf models (FreeTransfer-X)}. The FreeTransfer-X asks researchers to achieve CLT only with off-the-shelf English models but any labels, as formally defined in Section~\ref{sec:problem_definition}. 
To the best knowledge of the authors, it's the first time that the FreeTransfer-X is clearly defined.

To address the FreeTransfer-X, we propose a 2-step knowledge distillation~\citep[KD,][]{hinton2015distilling} framework based on mPLM, as shown in Figure~\ref{fig:overview}.
Given an off-the-shelf model $\theta_{src}$ in the source language (e.g. English), first we take $\theta_{src}$ as the teacher and an mPLM model $\theta_{src}^{m}$ as the student, then train $\theta_{src}^{m}$ on unlabeled corpus $\mathcal{D}_{src}$. Second, we take $\theta_{tgt}^{m}$ as the teacher and train a student $\theta_{tgt}$ on unlabeled corpus $\mathcal{D}_{tgt}$.
This cross-lingual transfer framework is label-free and applicable for any model architecture.
Experimental results demonstrate the effectiveness of the proposed framework on both sentence classification and sequence tagging. 

In short, the major contributions of this work include:
\begin{itemize}
    \item A novel cross-lingual transfer problem FreeTransfer-X is defined. The FreeTransfer-X asks researchers to achieve CLT from off-the-shelf models without using labels. It reduces the labeling cost and protects the labels in private domains such as medicine and business.
    \item We propose a 2-step knowledge distillation framework based on mPLMs, e.g. XLM-RoBERTa~\citep{xlmr-conneau-etal-2020-unsupervised}, to address the FreeTransfer-X. It significantly outperforms the NMT baselines on sentence classification and sequence tagging tasks. Besides, it's compatible with heterogeneous networks.
    \item Further analysis indicates abundant research potentials of the proposed framework. To improve the two distillation steps and the mPLM may benefit the framework.
\end{itemize}

\section{Methodology}\label{methodology}

\subsection{Problem Definition}\label{sec:problem_definition}
Denote the source language and the target language as $src$ and $tgt$ respectively.
Given an off-the-shelf model $\theta_{src}$ (e.g. English intent classifier), unlabeled in-domain corpus $\mathcal{D}_{src}$ and unlabeled in-domain corpus $\mathcal{D}_{tgt}$, the objective is to output a model $\theta_{tgt}$ in the target language $tgt$.
For simplicity in this paper, we constrain the target model $\theta_{tgt}$ to be of the same network architecture to the off-the-shelf source model $\theta_{src}$. 

\subsection{Basic Framework}
We propose to adopt knowledge distillation~\citep[KD,][]{hinton2015distilling} to address the FreeTransfer-X, since it can transfer knowledge from teacher models without knowing original labels.
In addition, knowledge distillation is free from network architectures and can be applied between heterogeneous networks, which benefits the deployment in various environment.
\subsubsection{Two-Step Knowledge Distillation} 
For a specific natural language processing (NLP) task, given a model $\theta_{src}$ and the unlabeled data $\mathcal{D}_{src}$ in the source language $src$ and the unlabeled data $\mathcal{D}_{tgt}$ in the target language $tgt$.
As shown in Figure~\ref{fig:overview}, we propose to train a model $\theta_{tgt}$ in the target language $tgt$ via 2 KD steps:
\begin{enumerate}
    \item Leverage the NLP capability of the off-the-shelf model $\theta_{src}$, e.g. an English sentence classifier $\theta_{en,cls}$. We distill knowledge from the teacher $\theta_{src}$ to the student mPLM $\theta_{src}^m$ on data $\mathcal{D}_{src}$.
    \item Due to the zero-shot cross-lingual transfer capability of the mPLMs, $\theta_{src}^m$ implicitly achieve the NLP capability on the target language $\theta_{tgt}^m$. Then similar to the step 1, we distill knowledge from the teacher $\theta_{tgt}^m$ to the student $\theta_{tgt}$ in the target language $tgt$ on data $\mathcal{D}_{tgt}$.
\end{enumerate}
The proposed framework works for arbitrary network including but not limited to Transformers~\citep{trm_NIPS2017_3f5ee243}, BiLSTM~\citep{bilstm} and CNN~\citep{kim-2014-convolutional}.

\subsubsection{Training Objectives}\label{sec:objectives}
The training is purely based on KD that no other training objectives is included.
We only apply KD between the classification distribution $P_{T}(\cdot)$ and $P_{S}(\cdot)$ of the teacher and the student respectively, which is compatible to arbitrary model architecture. 
Freezing the parameters of the teacher, we train the student by minimizing the Kullback-Leibler Divergence~\citep[$Div_{KL}$,][]{Joyce2011} between them.
Denote the prediction category as $\mathcal{C}=[c_0, c_1, ..., c_k]$, then the $Div_{KL}$ can be formalized as,

\begin{equation}\label{eq:KLDiv}
\begin{aligned}
    &Div_{KL}(P_{T}(\mathcal{C}|\cdot)\|P_{S}(\mathcal{C}|\cdot) \\
    &= \sum_{c_i\in\mathcal{\mathcal{C}}}P_{T}(c_i|\cdot)\log{(\frac{P_{T}(c_i|\cdot)}{P_{S}(c_i|\cdot)})}
\end{aligned}
\end{equation}

However, KD objectives of different NLU tasks varies a lot.
We classify NLU tasks into two categories: 1) sentence-level tasks like sentence classification, 2) word-level tasks like sequence tagging.
Given an input example $\mathcal{X}\in\mathcal{D}$ as a sequence of words $\mathcal{X}=[x_0, x_1, ..., x_n]$.
For sentence-level tasks, $\mathcal{X}$'s sentence-level category is $\mathcal{C_{\mathcal{X}}}$.
The teacher model and student model respectively output sentence-level prediction distribution $P_{T}(\mathcal{C_{X}}|\mathcal{X})$ and $P_{S}(\mathcal{C_{X}}|\mathcal{X})$.
For word-level tasks, $\mathcal{X}$'s word-level category is $\mathcal{C}_{x_{i}}, i\in[0,n]$.
Then the KD objective can be written as, 

\begin{equation}\label{eq:kd-loss-seq}
\begin{aligned}
    \mathcal{L} &= Div_{KL}(P_{T}(\mathcal{C}|\mathcal{X})\|P_{S}(\mathcal{C}|\mathcal{X})) \\
    &\text{where } \mathcal{C} = 
    \begin{cases}
        \mathcal{C_X} & , \text{sentence-level} \\
        \mathcal{C}_{x_i} & , \text{word-level}
    \end{cases}
\end{aligned}
\end{equation}

It's worth noting that word-level $Div_{KL}$ cannot be directly applied for heterogeneous teacher and student models since their tokenizations are different. 
In order to align the predictions of teacher and student, we only adopt the prediction on the first sub-word of each word.

\subsection{Enhanced Cross-Lingual Distillation}
To explore the potentials of improving the two KD steps, we propose to enhance them with machine translation (MT) and paraphrase generation (PG).

\subsubsection{Language Balanced Distillation} 
During the first KD step that training the mPLM from an English (i.e. the source language) classifier, to leverage the cross-lingual transferarability of mPLM, the conventional method is to train the mPLM only on the English corpus.
However, in our preliminary experiments, we notice that the mPLM's accuracy gap between English and the target languages are very huge. 
It's over 5\% between the English target model (94.0) and the average of all target models (88.4), as reported by \textit{2-step KD} in Table~\ref{tab:cls_matis_full_unlabeled}, Appendix~\ref{app:lang_wise_results}.

\begin{figure}[!h]
    \centering
    \includegraphics[width=1.\columnwidth]{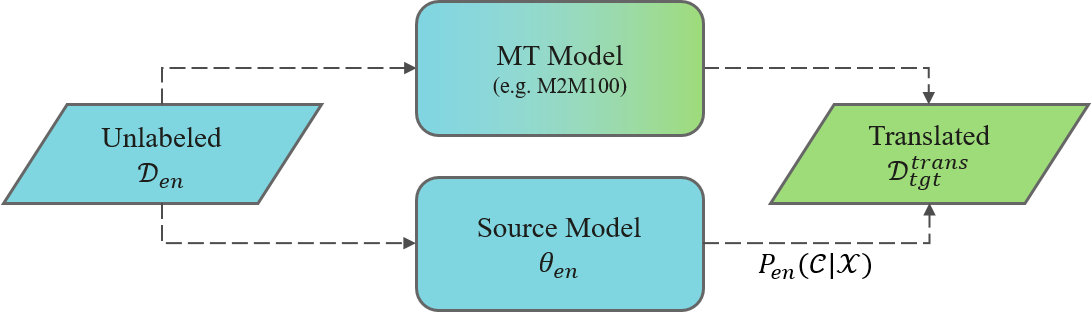}
    \caption{Language balanced distillation. Leverage the MT model to translate unlabeled English $\mathcal{D}_{en}$ into target languages $\mathcal{D}^{trans}_{tgt}$. Perform KD on the translated $\mathcal{D}^{trans}_{tgt}$ with $\theta_{en}$'s predicted distribution $P_{en}(\mathcal{C}|\mathcal{X})$.}
    \label{fig:balanced_dist}
\end{figure}

Hence, we propose to translate the unlabeled English corpus $\mathcal{D}_{en}$ to target languages $\mathcal{D}^{trans}_{tgt}$, as depicted by Figure~\ref{fig:balanced_dist}.
Since $\mathcal{D}_{en}$ and $\mathcal{D}^{trans}_{tgt}$ are aligned, source English model's predicted distribution $P_{en}(\mathcal{C}|\mathcal{X})$ of $\mathcal{D}_{en}$ can be directly applied to $\mathcal{D}^{trans}_{tgt}$.
In this way, KD is able to be performed on not only the source language but also the target languages.

As shown in the lower left of Figure~\ref{fig:overview}, the translated $\mathcal{D}^{trans}_{tgt}$ is incorporated in the training of KD step one.

\subsubsection{Language-Specific Data Augmentation} 
Inspired by data augmentation for KD~\citep{jiao-etal-2020-tinybert} and multilingual paraphrase generation~\citep{DBLP:journals/corr/abs-1911-03597}, we augment the unlabeled target corpus $\mathcal{D}_{tgt}$ via paraphrasing.

\begin{figure}[!h]
    \centering
    \includegraphics[width=1.\columnwidth]{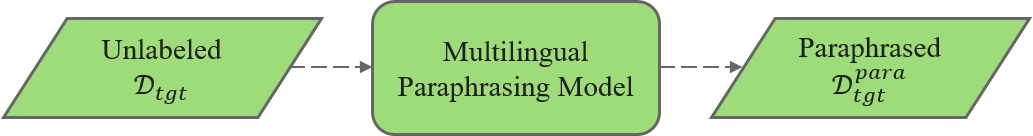}
    \caption{Language-specific data augmentation. We paraphrase the target corpus $\mathcal{D}_{tgt}$ into $\mathcal{D}^{para}_{tgt}$ as the augmented training data. KD is then performed on the mixture of $\mathcal{D}_{tgt}$ and $\mathcal{D}^{para}_{tgt}$.}
    \label{fig:data_augment}
\end{figure}

\section{Experiments}\label{sec:experiments}
\subsection{Datasets and Preprocessing}\label{sec:datasets_prep}
\noindent\textbf{MultiATIS++}~\citep{xu-etal-2020-end}
extends the Multilingual ATIS corpus~\citep{8461905} to 9 languages across 4 language families, including Indo-European (English, Spanish, German, French, Portuguese and Hindi), Sino-Tibetan (Chinese), Japonic (Japanese) and Altaic (Turkish).
It provides annotations for intent recognition (sentence classification) and slot filling (sequence tagging) for each languages.
The utterances are professionally translated from English and manually annotated.
MultiATIS++ includes 37,084 training examples and 7,859 testing examples.

\noindent\textbf{MTOP}~\citep{li-etal-2021-mtop} 
is a recently released multilingual NLU dataset covering 6 languages: English, German, French, Spanish, Hindi, Thai.
It's also manually annotated for intent recognition (sentence classification) and slot filling (sequence tagging).
MTOP provides a larger corpus consisting of 104,445 examples, of which 10\% is validation set and 20\% is testing set.

For each language, we randomly split both MultiATIS++ and MTOP into two balanced parts: \textit{annotated} and \textit{unannotated}.
The annotated parts are used to train and simulate the off-the-shelf source models while the unannotated parts are used for training the baselines and the proposed 2-step distillation model.
We tokenize Chinese, Japanese and Thai utterances using Jieba\footnote{https://github.com/fxsjy/jieba}, MeCab\footnote{https://github.com/polm/fugashi} and pythainlp\footnote{https://github.com/PyThaiNLP/pythainlp} respectively.

\subsection{Baselines}\label{sec:baselines}
\noindent\textbf{Translate-Test}~\citep{xnli-conneau-etal-2018}
is a machine translation based method. 
It performs two-pass inferences to tackle the FreeTransfer-X problem: 
1) translate the testing utterances into English (i.e. the source language) from the target language, 
2) predict on the translated English utterances with the off-the-shelf English model.

\noindent\textbf{Translate-Train-Pseudo}
is also based on machine translation.
It's a variant of the \textbf{Translate-Train}~\citep{xnli-conneau-etal-2018}, which translates English training examples into target languages and applies English annotations to the translated examples.
However, annotations are not provided in the FreeTransfer-X problem.
Hence, Translate-Train-Pseudo utilizes the prediction of the off-the-shelf English model to pseudoly annotates the translated examples.

\noindent\textbf{Gold-Supervised}
is for reference since it's trained with annotations.
It replaces the first distillation step of the proposed framework with gold-supervised training, in other words, the mPLM is supervised by gold annotations instead of the off-the-shelf English model.
It's supposed to be very strong.

\begin{table*}[ht]
    \centering
    \resizebox{2\columnwidth}{!}{
    \begin{tabular}{ll|ccc|ccc|c}
    \hline
    \multicolumn{2}{c|}{\multirow{2}{*}{Models}} & \multicolumn{3}{c|}{MTOP} & \multicolumn{3}{c|}{MultiATIS++} & \multirow{2}{*}{Avg} \\
    \multicolumn{2}{l|}{} & Transformer & BiLSTM & CNN & Transformer & BiLSTM & CNN & \\ \hline
    \multirow{2}{*}{Reference} & Off-the-shelf En source & 88.3 & 86.2 & 90.5 & 94.4 & 90.8 & 92.7 & 90.5 \\
                                & Gold-supervised target & 78.4 & 65.0 & 79.2 & 84.6 & 85.2 & 86.6 & 79.8 \\ \hline
    \multirow{2}{*}{Baselines} & Translate-test & 69.6 & 66.0 & 73.8 & 86.4 & 80.7 & 86.2 & 77.1 \\
                                & Translate-train-pseudo & 64.2 & 57.9 & 67.4 & 84.7 & 81.2 & 83.2 & 73.1 \\ \hline
    \multirow{3}{*}{Ours} & 2-step KD & 75.1 & 72.3 & 75.6 & 87.7 & 83.8 & 85.0 & 79.9 \\
                          & + Balanced distillation & 79.3 & 75.9 & 77.8 & \textbf{88.9} & 85.2 & 86.2 & 82.2 \\
                          & + Data augmentation & \textbf{79.6} & \textbf{79.1} & \textbf{78.8} & 88.7 & \textbf{86.4} & \textbf{86.9} & \textbf{83.3} \\ \hline
    \end{tabular}
    }
    \caption{Classification accuracy averaged over target languages. MTOP: de, es, fr, hi, th. MultiATIS++: de, es, fr, hi, ja, pt, tr, zh. For simplicity, the architecture of a target model is identical to its corresponding English source model.}
    \label{tab:cls_total_results}
\end{table*}

\subsection{Experiment Settings}
\subsubsection{Model Architectures}
We experiment with three mainstream NLU model architectures to verify the universality of the proposed framework.
They are used as the backbones of the off-the-shelf models $\theta_{src}$ and the output models $\theta_{tgt}$ in target language.

\noindent\textbf{Transformer} encoder~\citep{trm_NIPS2017_3f5ee243} models input sequences fully with Attention mechanism.
We follow the language modeling method of BERT~\citep{devlin-etal-2019-bert}. 
We adopt absolute positional encoding. 
The contextual representation vector of the first word is used for sentence classification.
Sequence tagging is based on the contextual representation of each word.

\noindent\textbf{Bidirectional LSTM (BiLSTM)}~\citep{bilstm} models input sequences via leveraging two stacked LSTM layers respectively from backward and forward directions.
We take the representation vector of the last word for sentence classification.
Word-level representation is used for sequence tagging like Transformer.

\noindent\textbf{Convolutional Neural Networks (CNN)}~\citep{kim-2014-convolutional} encodes input sequences with CNN modules.
We adopt three kind of 1-D kernels with kernel size of 3, 4 and 5.
Output vectors from all kernels and channels are concatenated as the representation for sentence classification.
Dilated CNN~\citep{strubell-etal-2017-fast} is adopted for sequence tagging.

\subsubsection{Training Details}
English is regarded as the source language in all the experiments.
Off-the-shelf English models are trained on the hold-out annotated English corpus as described in Section~\ref{sec:datasets_prep}.
All the experimented models are controlled in comparable model scale. 
AdamW~\citep{loshchilov2018decoupled} is adopted as the optimizer with $\epsilon=1e-8$.
We train the models for 50 epochs and take the checkpoint of the best validation accuracy as the final model.
Table~\ref{tab:num_params_archs} reports the hyper-parameters of the model architectures.

\begin{table}[!h]
    \centering
    \resizebox{.95\columnwidth}{!}{
    \begin{tabular}{l|cccc}
        \hline
         Model & Embed size & Hidden size & \#Layers & \#Params \\
        \hline
         Transformer & 256 & 256 & 4 & 5.3M \\
         BiLSTM & 256 & 512 & 2 & 5.3M \\
         CNN & 256 & 768 & 2 & 5.0M \\
        \hline
    \end{tabular}
    }
    \caption{Hyper-parameters of the experimented models.}
    \label{tab:num_params_archs}
\end{table}

Initial learning rate is decided based on a gradient-based searching heuristics proposed by~\citet{DBLP:journals/corr/Smith15a}, since in our preliminary experiments~\citet{DBLP:journals/corr/Smith15a} stably finds better learning rates than manual searching.
We build vocabularies of 10k words for each language via Byte Pair Encoding (BPE)~\citet{sennrich-etal-2016-neural}.
Experiments are implemented with PyTorch~\citep{NEURIPS2019_bdbca288} and conducted on a single Nvidia V100 32GB GPU.

\subsubsection{Auxiliary Models}
\noindent\textbf{M2M-100}~\citep{JMLR:v22:20-1307} is adopted as the MT system in our experiments. 
We apply the 418M model checkpoint from Huggingface\footnote{https://huggingface.co/facebook/m2m100\_418M}.

\noindent\textbf{XLM-RoBERTa}~\citep{xlmr-conneau-etal-2020-unsupervised} is adopted as the mPLM in the proposed 2-step distillation framework.

\subsection{Results}
Average accuracy across languages and models is given in Table~\ref{tab:cls_total_results} and Table~\ref{tab:tag_total_results}.
Language-wise results are provided in Appendix~\ref{app:lang_wise_results}.
\subsubsection{Sentence Classification}
%

\begin{table*}[ht]
    \centering
    \resizebox{2\columnwidth}{!}{
    \begin{tabular}{ll|ccc|ccc|c}
    \hline
    \multicolumn{2}{c|}{\multirow{2}{*}{Models}} & \multicolumn{3}{c|}{MTOP} & \multicolumn{3}{c|}{MultiATIS++} & \multirow{2}{*}{Avg} \\
    \multicolumn{2}{l|}{} & Transformer & BiLSTM & CNN & Transformer & BiLSTM & CNN & \\ \hline
    \multirow{2}{*}{Reference} & Off-the-shelf En source & 74.8 & 81.1 & 72.1 & 88.4 & 94.0 & 89.1 & 83.3 \\
                                & Gold-supervised target & 64.6 & 68.6 & 63.3 & 71.5 & 76.5 & 74.1 & 69.8 \\ \hline
    \multirow{1}{*}{Baselines} & Translate-test & 37.2 & 41.4 & 34.2 & 24.8 & 38.8 & 40.8 & 36.2 \\
                                & Translate-train-pseudo & 34.4 & 40.4 & 28.6 & 53.9 & 63.1 & 61.8 & 47.0 \\ \hline
    \multirow{1}{*}{Ours} & 2-step KD & \textbf{63.7} & \textbf{67.6} & \textbf{55.7} & \textbf{71.7} & \textbf{76.9} & \textbf{73.5} & \textbf{68.2} \\ \hline
    \end{tabular}
    }
    \caption{Sequence tagging F1 score averaged over target languages. MTOP: de, es, fr, hi, th. MultiATIS++: de, es, fr, hi, ja, pt, tr, zh. For simplicity, the architecture of a target model is identical to its corresponding English source model.}
    \label{tab:tag_total_results}
\end{table*}

\begin{table*}[ht]
    \centering
    \resizebox{.95\textwidth}{!}{
    \begin{tabular}{ll|c|c|cccccccc|c}
    \hline
    \multicolumn{2}{c|}{\multirow{2}{*}{Models}} & Original & Finetuned & \multicolumn{9}{c}{Zero-Shot Cross-Lingual Transfer}          \\
    \multicolumn{2}{c|}{} & en & en & de & es & fr & hi & \textbf{ja} & pt & \textbf{tr} & \textbf{zh} & Avg \\ \hline
    \multicolumn{2}{l|}{Gold-supervised} & - & 97.9 & 97.6 & 97.4 & 97.4 & 92.4 & 90.6 & 97.3 & 83.8 & 92.8 & 93.7 \\ \hline
    \multirow{2}{*}{Transformer} & Naive KD & \multirow{2}{*}{94.4} & 97.2 & 96.9 & 96.8 & 96.2 & 90.8 & 90.1 & 95.6 & 84.3 & 91.3 & 92.8 \\
    & + Balanced distillation & & 97.5 & 97.6 & 96.9 & 96.6 & 95.4 & 96.3 & 96.1 & 90.8 & 97.5 & 95.9\\ \hline
    \multirow{2}{*}{BiLSTM} 
    & Naive KD & \multirow{2}{*}{90.8} & 93.2 & 93.5 & 93.5 & 93.2 & 90.4 & 83.4 & 93.5 & 77.2 & 85.0 & 88.7 \\
    & + Balanced distillation & & 92.4 & 93.3 & 93.7 & 92.4 & 91.8 & 91.9 & 93.1 & 86.6 & 92.8 & 92.0 \\ \hline
    \multirow{2}{*}{CNN} 
    & Naive KD & \multirow{2}{*}{92.7} & 94.7 & 91.8 & 94.1 & 93.2 & 90.3 & 90.9 & 94.2 & 83.5 & 90.3 & 91.0 \\
    & + Balanced distillation & & 92.8 & 93.2 & 92.9 & 92.4 & 91.8 & 91.6 & 93.2 & 89.2 & 92.9 & 92.2 \\ \hline
    \end{tabular}
    }
    \caption{Classification accuracy of the finetuned mPLM models, i.e. XLM-RoBERTa. Evaluated on MultiATIS++. \textit{Gold-supervised} is trained with gold annotations. \textbf{Bold languages} is not in the Indo-European language family as English.}
    \label{tab:dist_step_1}
\end{table*}

As shown in Table~\ref{tab:cls_total_results}, the proposed 2-step KD framework significantly outperforms the MT baselines on most model architectures, except for the CNN of \textit{Translate-test}.
Although \textit{Translate-test} is strong in a very few cases, it requires 2-pass inference (MT and classification) that results in a high latency. 
On the contrary, the proposed framework directly produces classification models in the target languages, which is more efficient.
In addition, the \textit{language-balanced distillation} and \textit{language-specific data augmentation} further enhance our model to a large extent, +2.3\% and +1.1\% respectively.
Language-wise results in Table~\ref{tab:cls_matis_full_unlabeled} demonstrate the robustness of our method across various languages.

To our surprise, the naive \textit{2-step KD} model even performs on par with the \textit{Gold-supervised} reference on average.
We guess it's due to the regularization effects of knowledge distillation that brings a good generalizability to the proposed model.
It implies the proposed framework may be a annotation-free alternative to current zero-shot cross-lingual transfer framework.

However, comparing the results of the English source model and those of the target models in Table~\ref{tab:cls_matis_full_unlabeled}, the cross-lingual transferred models still lag far behind the original English models.
There is a great potential of the proposed framework.

\subsubsection{Sequence Tagging}
%

On the sequence tagging task, the proposed model beats the baselines by a wide margin.
The MT-based baselines perform very poor on this task due to the error from word-level annotation alignment.
Also because of the alignment error, we do not apply \textit{language balanced distillation} and \textit{language-specific data augmentation} on this task.

As to the comparison with the \textit{Gold-supervised} reference, our model performs slightly worse than it. 
It may due to the insufficient knowledge distillation from the teacher to the student, which comes from the discrepancy between teacher's and student's tokenizations.
Although, as described in Section~\ref{sec:objectives}, we perfectly align their prediction at word-level, only the first subword of each word is used for distillation.
More informative subword-level aligning and distillation methods can be explored.
We leave this problem for the future research.
Besides, similar to sentence classification, gap between the English source model and the transferred target models is huge, as shown in Table~\ref{tab:tag_total_results}.

In sum, both experimental results on sentence classification and sequence tagging demonstrate that the proposed model is significantly stronger than MT-based cross-lingual transfer methods.
Furthermore, the proposed model only slightly lags behind or even performs on par with the strong \textit{Gold-supervised} reference, which is not able to address the FreeTransfer-X problem.

\section{Further Analysis}
In order to explore the potential of the proposed framework, we analyze it in more details.
For simplicity, experiments in this Section are conducted only on the MultiATIS++ sentence classification task.
\subsection{Effects of the Distillation}\label{sec:effect_dist} 
\begin{table*}[]
    \centering
    \resizebox{.95\textwidth}{!}{
    \begin{tabular}{ll|c|c|cccccccc|c}
    \hline
    \multicolumn{2}{c|}{\multirow{2}{*}{Models}} & Original & Finetuned & \multicolumn{9}{c}{Zero-Shot Cross-Lingual Transfer}          \\
    \multicolumn{2}{c|}{} & en & en & de & es & fr & hi & ja & pt & tr & zh & Avg \\ \hline
    \multirow{2}{*}{Transformer} & XLM-RoBERTa & \multirow{2}{*}{94.4} & 97.2 & 96.9 & 96.8 & 96.2 & 90.8 & 90.1 & 95.6 & 84.3 & 91.3 & 92.8 \\
    & mBERT & & 96.9 & 88.4 & 92.4 & 93.8 & 81.1 & 85.7 & 94.0 & 73.7 & 83.2 & 86.5 \\ \hline
    \multirow{2}{*}{BiLSTM}
    & XLM-RoBERTa & \multirow{2}{*}{90.8} & 93.2 & 93.5 & 93.5 & 93.2 & 90.4 & 83.4 & 93.5 & 77.2 & 85.0 & 88.7 \\
    & mBERT & & 92.3 & 80.4 & 87.5 & 82.5 & 79.2 & 79.4 & 82.3 & 76.5 & 75.0 & 80.3 \\ \hline
    \multirow{2}{*}{CNN}
    & XLM-RoBERTa & \multirow{2}{*}{92.7} & 94.7 & 91.8 & 94.1 & 93.2 & 90.3 & 90.9 & 94.2 & 83.5 & 90.3 & 91.0 \\
    & mBERT & & 93.3 & 82.6 & 86.9 & 87.9 & 78.1 & 78.7 & 88.2 & 72.9 & 80.9 & 82.0 \\ \hline
    \end{tabular}
    }
    \caption{Classification accuracy of XLM-RoBERTa and mBERT. \textit{Step-1 KD}: off-the-shelf English model -> mPLM. The mPLMs are finetuned and evaluated on MultiATIS++.}
    \label{tab:mbert_step_1}
\end{table*}

\begin{table*}[]
    \centering
    \resizebox{.95\textwidth}{!}{
    \begin{tabular}{ll|c|c|cccccccc|cc}
    \hline
    \multicolumn{2}{c|}{\multirow{2}{*}{Models}} & Original & Transferred & \multicolumn{10}{c}{Transferred Target Languages}          \\
    \multicolumn{2}{c|}{} & en & en & de & es & fr & hi & ja & pt & tr & zh & Avg & $\Delta$ \\ \hline
    \multirow{2}{*}{Transformer} & XLM-RoBERTa & \multirow{2}{*}{94.4} & 94.0 & 94.4 & 93.3 & 90.8 & 85.7 & 81.3 & 92.4 & 76.5 & 87.1 & 87.7 & -5.1\\
    & mBERT & & 95.6 & 86.2 & 91.8 & 92.8 & 79.8 & 81.7 & 90.9 & 72.9 & 80.3 & 84.6 & -1.9 \\ \hline
    \multirow{2}{*}{BiLSTM}
    & XLM-RoBERTa & \multirow{2}{*}{90.8} & 89.8 & 89.8 & 89.7 & 89.5 & 84.1 & 78.1 & 84.9 & 71.5 & 83.1 & 83.8 & -4.9 \\
    & mBERT & & 89.6 & 80.9 & 82.9 & 81.0 & 78.5 & 76.0 & 83.7 & 71.7 & 78.6 & 79.2 & -1.1 \\ \hline
    \multirow{2}{*}{CNN}
    & XLM-RoBERTa & \multirow{2}{*}{92.7} & 90.7 & 87.9 & 88.1 & 87.8 & 85.0 & 83.2 & 86.3 & 75.2 & 86.7 & 85.0 & -6.0 \\
    & mBERT & & 89.1 & 79.4 & 82.9 & 81.5 & 76.0 & 78.2 & 82.4 & 72.0 & 79.5 & 79.0 & -3.0 \\ \hline
    \end{tabular}
    }
    \caption{Classification accuracy of the target models, distilled from XLM-RoBERTa and mBERT respectively. \textit{Step-2 KD}: mPLM -> target model. \textit{$\Delta$}: changes w.r.t Table~\ref{tab:mbert_step_1}. The target models are transferred and evaluated on MultiATIS++.}
    \label{tab:mbert_step_2}
\end{table*}

Table~\ref{tab:dist_step_1} reports the accuracy of the XLM-RoBERTa finetuned from gold annotations, Transformer teacher, BiLSTM teacher and CNN teacher.

First, compare the \textit{Original} with the \textit{Naive KD Finetuned} of each model respectively.
It's very interesting that the accuracy of the student mPLM is consistently higher than its teacher.
The XLM-RoBERTa students gain 2.8\%, 2.4\% and 2.0\% improvement from the \textit{Original} teachers as Transformer, BiLSTM and CNN respectively.
The phenomenon implies the general effectiveness of language modeling of mPLMs.
We conjecture the improvement comes from two aspects:
1) mPLMs' generalizability learn from the large-scale pre-training,
2) the large model scale of mPLMs, which enhances its NLU capability.
Besides, the improvement with respect to the \textit{Original} varies across model architectures.
Especially when compare Transformer (+2.8\%) to CNN (+2.0\%), although the Transformer's student XLM-RoBERTa performs much closer to the \textit{Gold-supervised}, it still improves greater than the CNN's student.
Since the XLM-RoBERTa is Transformer-based network, it implies that the knowledge distillation performs better if the architectures of the teacher and the student are more similar.

Second, under the cross-lingual transfer condition, although the \textit{Gold-supervised} outperforms the \textit{Naive KD} on most target languages, it performs weaker on Turkish (tr).
It demonstrates the better generalizability and few-shot performance of the \textit{Naive KD}, since Turkish is a low-resource language in MultiATIS++.
The number of training examples of Turkish (578) is less than other languages (4488).

Third, the effectiveness of the proposed \textit{language balanced distillation} is very clear. 
In the comparison between the \textit{Naive KD} and \textit{+ Balanced distillation}, the accuracy is highly boosted almost on all the target languages.
This improvement is particularly significant on the languages that is not in the same family of English: Hindi (hi), Japanese (ja), Turkish (tr) and Chinese (zh).
A future research topic is to improve \textit{language balanced distillation} on the languages similar to the source language, e.g. European languages to English.
Data selection algorithms may have potentials.

In sum, the proposed framework and distillation method is effective and of strong generalizability. 
Future researches on heterogeneous distillation and data selection may benefit the proposed framework.

%

\subsection{Effects of mPLM Models}\label{sec:effect_mplm}
Table~\ref{tab:mbert_step_1} and Table~\ref{tab:mbert_step_2} respectively reports accuracy of the step-1 KD and step-2 KD in the proposed framework.
According to Table~\ref{tab:mbert_step_2}, the choice of mPLM is critical to the target models' performance.
Performance with XLM-RoBERTa as the mPLM is stronger than with mBERT.
However, there are interesting observations we should notice.

First, observe the performance changes ($\Delta$) of the \textit{Step-2 KD}: from the mPLM teacher to the target model student. 
We notice that the performance drop of mBERT is slighter than the XLM-RoBERTa's, based on the results of the average score in Table~\ref{tab:mbert_step_2} minus those in Table~\ref{tab:mbert_step_1}.
It implies that as the capability of mPLM increases, the KD dissipation tends to increase as well.
Similar to the analysis in Section~\ref{sec:effect_dist}, the KD dissipation may come from:
1) the pre-trained language model that the target models lack of,
2) discrepancy between the model size of the mPLM and the target models.
Hence, performance based on XLM-RoBERTa drops more due to its gap to the target models is greater than mBERT's in both the two aspects of discrepancy.
To reduce the KD dissipation, researches should focus on how to reduce the model discrepancy between mPLM and the target model, e.g. improve the language modeling capability of the target model.
Besides, the performance difference among model architectures is consistent, either based on XLM-RoBERTa or mBERT.
It further evidences that the proposed framework is general and works well for different model architectures.

\subsection{Cross-Architecture Transfer}
To analyze the proposed framework in a more general setting, we free the architecture ties of the off-the-shelf English models and the target models to be heterogeneous, that the source and target models can be different.

\begin{figure}[!h]
    \centering
    \includegraphics[width=.8\columnwidth]{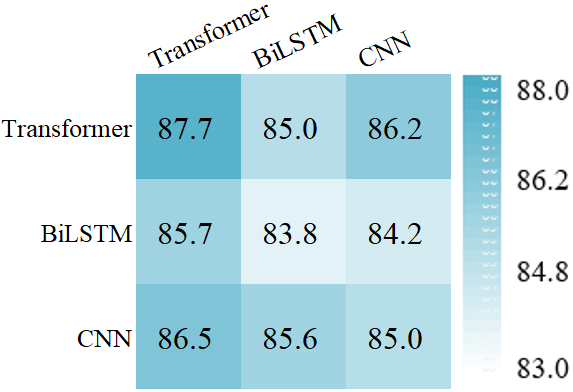}
    \caption{Classification accuracy of the target models via cross-architecture transfer, averaged over all target languages. Transfer from rows to columns. \textit{Row}: architectures of source English models, \textit{Column}: architectures of the target models. Experimented on MultiATIS++.}
    \label{fig:cross_arch_heatmap_abs}
\end{figure}

As depicted in Figure~\ref{fig:cross_arch_heatmap_abs}, the transfer performs the best when taking Transformer as both the source and target models.
The worst comes to the transfer between BiLSTM models.
On one side, the advantage of the Transformer architecture may be a reason.
On the other side, it reconfirms the observation that the more similar teacher and student models are, the better transfer performance comes.

Besides, taking the BiLSTM as the source or target model consistently result in lower accuracy, no matter what the corresponding target or source models are.
Hence, we guess the architecture similarity between BiLSTM and the Transformer-based mPLM is lower than that between CNN and the mPLM.
We leave this for future work.

\begin{figure}[!h]
    \centering
    \includegraphics[width=.8\columnwidth]{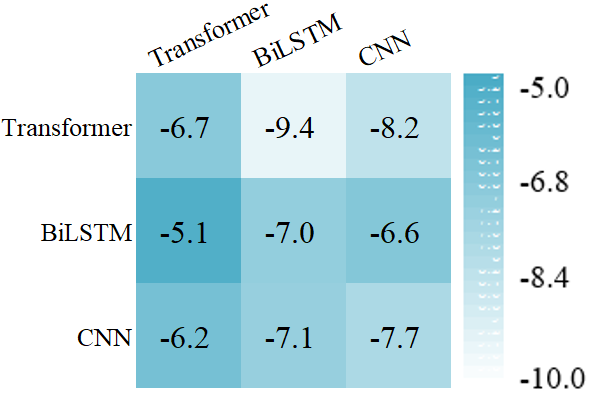}
    \caption{Accuracy drop from the source English models to the target models, averaged over all target languages. Transfer from rows to columns. \textit{Row}: architectures of source English models, \textit{Column}: architectures of the target models. Experimented on MultiATIS++.}
    \label{fig:cross_arch_heatmap_rel}
\end{figure}

In addition, we study the accuracy drop from the source English models to the target models, as shown in Figure~\ref{fig:cross_arch_heatmap_rel}.
From the perspective of the source model, the drop is the least when BiLSTM is the source.
From the perspective of the target model, the drop is the least when Transformer is the target.
It reveals an asymmetry between the two KD steps with respect to the mPLM.
To reduce the KD dissipation to the largest extent, it seems mPLM should be distilled from a weaker teacher architecture (e.g. BiLSTM) before teaching a stronger student architecture (e.g. Transformer).

In brief, the proposed framework works for heterogeneous cross-lingual transfer.
The future work may focus on how to define the similarity between model architectures and how to evaluate the source-target model pairs.

\section{Conclusions}
In this paper, we define a novel cross-lingual transfer (CLT) problem - FreeTransfer-X, especially for CLT in private scenarios such as medical and business.
The FreeTransfer-X is defined to transfer knowledge from off-the-shelf models in rich-resource languages to poor-resource languages, without labeled corpora.
To address the FreeTransfer-X, we propose a 2-step knowledge distillation (2-step KD) framework based on multilingual pre-trained language models.
In addition, two data augmentation methods for cross-lingual KD are proposed to boost the performance of the 2-step KD framework.
Experimental results clearly demonstrate the effectiveness of the proposed framework.
It's worth noting that the proposed KD framework can be applied between heterogeneous models, which benefits the deployment in different environment.
Further analyses point out various research directions for future work.

\bibliography{anthology,preprint,urls}
\bibliographystyle{acl_natbib}

\clearpage

\appendix
\section{Language-Wise Results}\label{app:lang_wise_results}
Here we list the detailed language-wise experimental results of Table~\ref{tab:cls_total_results} and Table~\ref{tab:tag_total_results} for reference.

\begin{table*}[ht]
    \centering
    \resizebox{1.\textwidth}{!}{
    \begin{tabular}{lll|c|ccccccccc|c}
        \hline
        \multicolumn{3}{c|}{\multirow{2}{*}{Models}} & Source & \multicolumn{10}{c}{Targets} \\
         & & & en & en & de & es & fr & hi & ja & pt & tr & zh & Avg \\
        \hline
        \multicolumn{2}{c}{Reference} & Gold-supervised & - & 88.5 & 89.7 & 89.6 & 91.9 & 81.9 & 79.5 & 86.3 & 73.4 & 89.1 & 85.2 \\ 
        \hline
        \multirow{5}{*}{Transformer} & \multirow{2}{*}{Baselines} & Translate-test & \multirow{5}{*}{94.4} & 92.5 & 90.0 & 88.1 & 90.6 & 83.7 & 86.8 & 88.1 & 75.1 & 88.7 & 86.4 \\
        & & Translate-train-pseudo & & 92.7 & 89.4 & 90.0 & 90.5 & 83.1 & 74.8 & 90.1 & 74.1 & 85.9 & 84.7 \\
        \cline{2-3}\cline{5-14}
        & \multirow{3}{*}{Ours} & 2-step KD & & 94.0 & 94.4 & 93.3 & 90.8 & 85.7 & 81.3 & 92.4 & 76.5 & 87.1 & 87.7 \\
        & & + Balanced distillation & & 94.3 & 93.5 & 92.9 & 95.0 & 84.0 & 83.0 & 93.3 & 78.5 & 90.7 & 88.9 \\
        & & + Data augmentation & & 94.7 & 93.6 & 93.3 & 94.7 & 84.2 & 83.7 & 93.2 & 77.6 & 89.7 & 88.7 \\
        \hline
        \multirow{5}{*}{BiLSTM} & \multirow{2}{*}{Baselines} & Translate-test & \multirow{5}{*}{94.4} & 87.1 & 84.2 & 83.4 & 85.8 & 77.9 & 81.5 & 84.4 & 64.1 & 84.2 & 80.7 \\
        & & Translate-train-pseudo & & 87.8 & 85.0 & 85.4 & 86.9 & 80.4 & 73.5 & 86.1 & 72.0 & 80.4 & 81.2 \\
        \cline{2-3}\cline{5-14}
        & \multirow{3}{*}{Ours} & 2-step KD & & 89.8 & 89.8 & 89.7 & 89.5 & 84.1 & 78.1 & 84.9 & 71.5 & 83.1 & 83.8 \\
        & & + Balanced distillation & & 90.7 & 89.4 & 88.7 & 88.6 & 82.5 & 81.7 & 86.1 & 76.2 & 88.6 & 85.2 \\
        & & + Data augmentation & & 89.1 & 90.6 & 90.9 & 88.7 & 83.4 & 82.5 & 86.7 & 77.9 & 90.3 & 86.4 \\
        \hline
        \multirow{5}{*}{CNN} & \multirow{2}{*}{Baselines} & Translate-test & \multirow{5}{*}{94.4} & 90.7 & 86.6 & 86.3 & 88.7 & 86.9 & 85.3 & 88.2 & 81.0 & 86.3 & 86.2 \\
        & & Translate-train-pseudo & & 86.3 & 84.4 & 85.6 & 86.5 & 83.8 & 79.8 & 82.2 & 77.5 & 85.9 & 83.2 \\
        \cline{2-3}\cline{5-14}
        & \multirow{3}{*}{Ours} & 2-step KD & & 90.7 & 87.9 & 88.1 & 87.8 & 85.0 & 83.2 & 86.3 & 75.2 & 86.7 & 85.0 \\
        & & + Balanced distillation & & 89.1 & 88.5 & 87.8 & 88.8 & 85.7 & 83.0 & 86.2 & 79.2 & 90.6 & 86.2 \\
        & & + Data augmentation & & 89.6 & 89.8 & 89.2 & 89.0 & 86.0 & 83.8 & 87.1 & 79.4 & 90.9 & 86.9 \\
        \hline
    \end{tabular}
    }
    \caption{Sentence classification accuracy on MultiATIS++.}
    \label{tab:cls_matis_full_unlabeled}
\end{table*}


\end{document}